\documentclass[lettersize,journal]{IEEEtran}
\usepackage{amsmath,amsfonts}
\usepackage[ruled,vlined]{algorithm2e}
\usepackage{array}
\usepackage{textcomp}
\usepackage{stfloats}
\usepackage{url}
\usepackage{verbatim}
\usepackage{graphicx}
\usepackage{caption}
\usepackage{subcaption}
\usepackage{cite}
\usepackage{booktabs}

\newcolumntype{M}[1]{>{\centering\arraybackslash}m{#1}}
\begin{document}

\title{Bin-picking of novel objects through category-agnostic-segmentation: RGB matters}

\author{Prem Raj$^1$, Sachin Bhadang$^1$, Gaurav Chaudhary$^1$, Laxmidhar Behera$^{1,2}$ ,  Tushar Sandhan$^1$
\thanks{$^1$
		Intelligence Systems and Control Lab \textit{Indian Institute of Technology} Kanpur, India \{praj, sachinb20, gauravch, lbehera, sandhan\}@iitk.ac.in}
	\thanks{$^2$
		\textit{Indian Institute of Technology} Mandi, India director@iitmandi.ac.in}
}



\maketitle

\begin{abstract}
This paper addresses category-agnostic instance segmentation for robotic manipulation, focusing on segmenting objects independent of their class to enable versatile applications like bin-picking in dynamic environments. Existing methods often lack generalizability and object-specific information, leading to grasp failures. We present a novel approach leveraging object-centric instance segmentation and simulation-based training for effective transfer to real-world scenarios. Notably, our strategy overcomes challenges posed by noisy depth sensors, enhancing the reliability of learning. Our solution accommodates transparent and semi-transparent objects which are historically difficult for depth-based grasping methods. Contributions include domain randomization for successful transfer, our collected dataset for warehouse applications, and an integrated framework for efficient bin-picking. Our trained instance segmentation model achieves state-of-the-art performance over \textit{WISDOM} public benchmark~\cite{danielczuk2019segmenting} and also over the custom-created dataset. In a real-world challenging bin-picking setup our bin-picking framework method achieves $98\%$ accuracy for opaque objects and $97\%$ accuracy for non-opaque objects, outperforming the state-of-the-art baselines with a greater margin.
\end{abstract}


\section{Introduction}
Category-agnostic instance segmentation is the technique to segment the individual objects in the scene regardless of their class \cite{danielczuk2019segmenting}. This method can be utilized for various robotic manipulation applications, such as robotic bin-picking of novel objects. Prominently, the instance-segmentation problem had been studied for cases with predefined semantic classes \cite{segmentation_traditional1,segmentation_traditional2}. This might be useful in bin-picking for a limited type of known object. However, this is not feasible for bin-picking in practical scenarios such as warehouse automation where new types of objects are introduced regularly.

There are various state-of-the-art solutions exist for bin-picking of novel diverse objects that directly predict the optimal grasp pose without a pre-segmentation step~\cite{dexnet2, dexnet4, case_baseline, graspfusionnet, suction_2022uncertainty}. They have certain disadvantages. First, these are gripper-centric solutions, and hence, a solution designed for one type of gripper (e.g. a parallel jaw gripper) can not be easily extendable to other types of gripper (i.e. a suction gripper). The category-agnostic instance segmentation is the object-centric solution and thus can be easily extended to any type of gripper. Secondly, the direct grasp-pose prediction methods do not have any object-specific information, and the predicted optimal grasp-pose might result in a grasp failure during the grasp attempt for various reasons (e.g. object slips when grasped from the corner side rather than grasping from its middle \cite{case_baseline}).

\begin{figure}[t]
	\centering
	\includegraphics[scale=0.75]{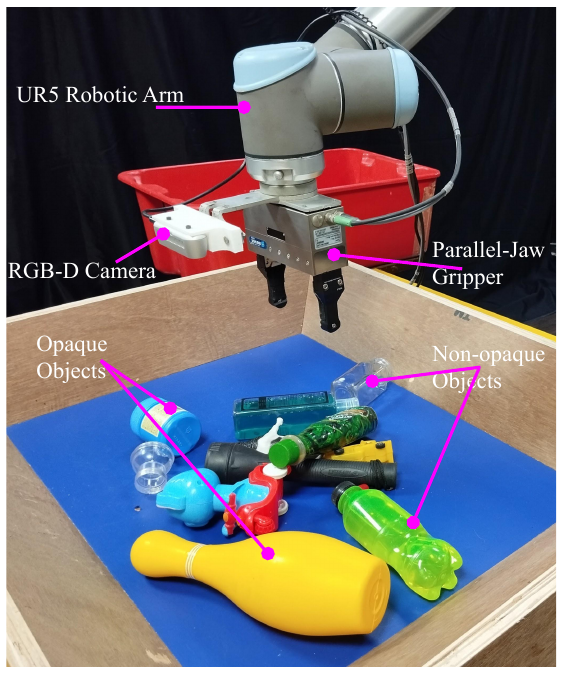}
	\caption{Experimental setup for our bin-picking problem with unknown diverse objects. The objects consist of opaque and non-opaque (i.e. transparent and translucent) objects. Depth sensing technology has become cheaper and easily available to common users. However, commodity-level sensors often generate noisy depth maps, especially for non-opaque objects. Our proposed method can grasp opaque and non-opaque objects in the bin, reliably, even in such a challenging scenario.}
	\label{fig:intro}
	\vspace{-3ex}
\end{figure}

Convolutional neural networks (CNN) modules are being used by the state-of-the-art solutions for the instance-segmentation tasks. However, the availability of labeled training data remains a major challenge as the process of data labeling is labor-intensive and costlier \cite{coco}. To overcome this, many solutions have used simulations for auto-generating the training data \cite{dexnet2, raj2020learning} followed by sim-to-real transfer of the learning for real-world deployment \cite{sim2real}. Previously, It was believed that real-world visuals differ significantly from the simulated world and hence the sim-to-real transfer is not promising in the case of models trained with only synthetic RGB data \cite{baseline_corl20, hofer2021sim2real}. Subsequently, leading work for learning class agnostic segmentation for bin-picking has shown that if the CNN module is trained over only the simulated depth images, the learning can directly be employed in the real world \cite{danielczuk2019segmenting}. Recently, many similar followed works~\cite{baseline_corl20, baseline_corl21, baseline_tro21, baseline_icra22} have fused RGB features with depth features and shown improved results in this context. However, these methods are tested in the real world with high-quality industrial-grade costlier depth sensors that produce very accurate depth maps with high precision. As verified in our work, this direct transfer of learning with depth-map inputs does not work for noisy depth maps produced with low-cost depth sensors, such as Realsense D435i, that are widely used in the research community in general. The trained network is found to be highly sensitive to noise in the depth maps.

One recent work~\cite{cas_close_baseline} talks about this issue and purposed to augment the simulated depth maps with manually modeled noise profiles to mimic the real-world noise. However, the modeling of noise is camera-specific and does not provide a generalized solution. In our work, we re-purpose the problem of category-agnostic instance segmentation in the case of not-so-high-precision depth sensors and show that a model trained with simulated color (RGB) images can directly transfer to the real world with performance to the level of the state-of-the-art if a carefully designed domain randomization strategy~\cite{domain-randomization} is used.  
Additionally, our method can segment effectively the transparent and the semi-transparent objects, enabling them to be grasped with ease, which always has been a great challenge for depth input modality-based methods as depth sensing is poorer for such objects \cite{dexnet4, case_baseline}.
The effectiveness of the proposed method has also been shown by performing real-world bin-picking trials in a challenging bin-picking setup. The details are further elaborated in Section \ref{sec:method}. 

In summary, the main contributions of our work are as follows:
\begin{itemize}
    \item Re-purposing the sim-to-real transfer of category-agnostic instance segmentation learning amidst the noisy depth sensing.
    \item A method to generate simulated training samples with domain randomization for sim-to-real transfer with RGB images.
    \item A simulated as well as a real dataset for category-agnostic instance segmentation in the context of warehouse applications for training and evaluation purposes.
    \item An integrated bin-picking framework that can also grasp transparent and semi-transparent objects effectively. The framework uses the purposed instance-segmentation method and an analytical grasp evaluation method \cite{case_baseline}.
\end{itemize}

\section{Related Works}
Broadly, our work comprises instance segmentation, sim-to-real transfer, and bin picking of diverse objects. To gain a better understanding of the existing research in these areas, we will comprehensively review related works in each of these categories.

\subsection{Instance Segmentation}
Instance segmentation, which involves simultaneously detecting objects and segmenting them into pixel-level masks, has gained significant attention in recent years due to its practical applications in autonomous driving, robotics, and medical imaging. Mask R-CNN \cite{he2017mask}, one of the most widely used instance segmentation methods, extends the popular object detection framework, Faster R-CNN \cite{ren2015faster}, by adding a segmentation branch that predicts the object mask in parallel with the object classification and bounding box regression tasks. Building on top of Mask R-CNN, many recent works have aimed to improve the accuracy and efficiency of instance segmentation, including the use of advanced backbones such as shufflenet~\cite{zhang2018shufflenet}, feature pyramid networks \cite{fpn}, and efficient training strategies \cite{chen2021big}. Another important area of research in instance segmentation is panoptic segmentation, which combines instance segmentation with semantic segmentation to provide a unified view of the scene \cite{panoptic}.

Category-agnostic instance segmentation detects and segments all object instances in an image, without prior knowledge of object categories. This technique is promising for robotics applications, as it enables robust perception and interaction in unstructured environments, where there are no predefined categories of objects \cite{cas1-mousavian2018joint, cas5-zhou2020object}. It has the potential for complex tasks, such as bin-picking \cite{cas2-li2020real, cas3-qin2019efficient, cas4-yin2021bin, danielczuk2019segmenting}, that require object localization in cluttered settings.

\subsection{Sim-to-real transfer}
Sim-to-real transfer is an important research area in robotics that focuses on developing techniques to transfer machine learning models trained in simulation to real-world settings. A variety of methods have been proposed for sim-to-real transfer, including domain randomization \cite{domain-randomization, james2018transferring}, data augmentation \cite{augmentation2019survey}, and adversarial training \cite{james2019sim}. Recent works have explored the use of sim-to-real transfer in a range of applications, such as robot grasping \cite{dexnet4}, navigation \cite{navi2018sim}, motion planning \cite{raj2020learning} and locomotion \cite{loco2018deep}.
In the case of the bin-picking problem, recently there are some works \cite{dexnet4, danielczuk2019segmenting, depth_transfer_icra_17} that have shown that CNN models trained purely over synthetic depth maps can be directly transferred to the real world. However, in contrast to these findings, we have found that this is only true in the case of high-precision noise-free depth sensing which is costlier. Instead, models trained over only RGB images with appropriate domain randomization can successfully transfer the learning to the real world without any further finetuning. The not-so-perfect depth maps from the low-cost depth sensors can still be useful for subsequent steps in the bin-picking such as grasp pose evaluation \cite{case_baseline}.

\subsection{Bin-picking}
The robotic bin-picking problem has a wide range of formulations depending on the type of objects in the bin (homogeneous or heterogeneous), the target application (warehouse automation or industrial parts handling), and the perception system (camera types, camera positioning, etc.) \cite{bin-picking-survey-2020, bin-picking-survey-2}. Specifically, in this paper, we are focusing on the bin-picking solutions that have a use-case in a warehouse automation application where a large number of novel objects with different shapes, sizes, colors, and textures, need to be handled \cite{bin-picking-survey-2020}. One type of solution for this category of bin-picking is designed to be gripper-specific \cite{dexnet2, dexnet4, antipodal, baseline, levine2018learning, graspfusionnet, raj2022domain, raj_tase23}. On the other hand, the gripper-agnostic works in this category involve an object-centric approach \cite{danielczuk2019segmenting}. One type of object-centric approach uses 3D CAD model of the target object \cite{CAD-bin-picking} which is not suitable for novel objects whose CAD model is not available. Another type of solution in this category is to apply a category-agnostic instance segmentation \cite{danielczuk2019segmenting, baseline_corl20, baseline_icra22, baseline_tro21, baseline_corl21, cas_close_baseline} which is also the focus of our purposed approach. Our work is most closely related to \cite{danielczuk2019segmenting} which is category-agnostic instance segmentation for bin-picking of unknown novel objects via sim-to-real transfer. 


\begin{figure*}[t]
    \centering

    \begin{subfigure}[b]{0.9\textwidth}
       \includegraphics[width=1\linewidth]{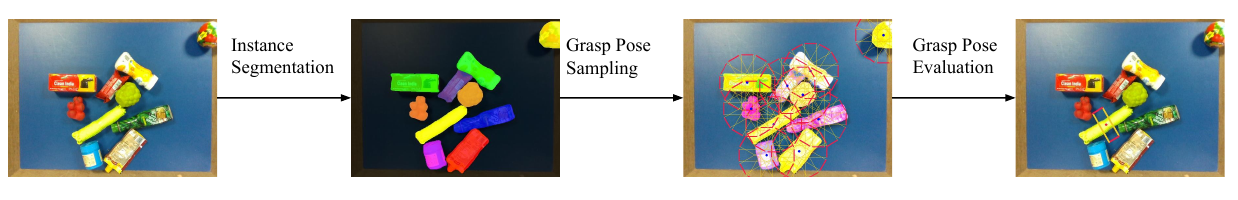}
       \caption{Whole bin-picking pipeline.}
       \label{fig:md1} 
    \end{subfigure}

    \begin{subfigure}[b]{0.9\textwidth}
       \includegraphics[width=1\linewidth]{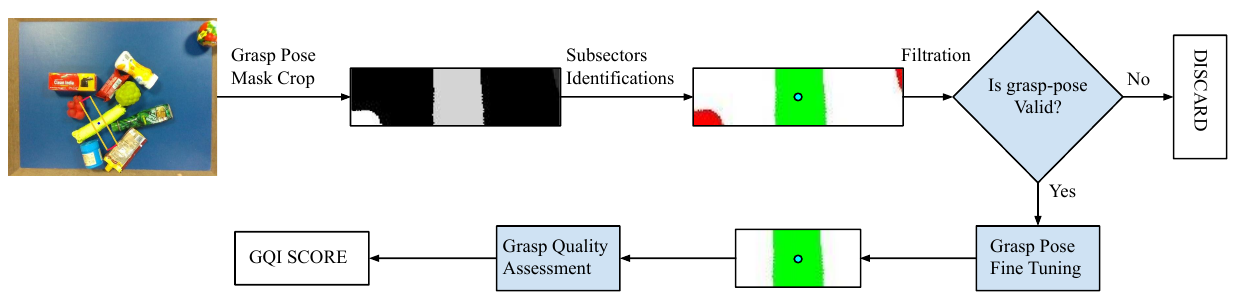}
       \caption{Grasp-Pose Evaluation process (depicted for a single grasp pose).}
       \label{fig:md2} 
    \end{subfigure}
   
    \caption{Schematic diagram of our proposed method (Best viewed in the color). Part (a) shows the overview of the entire end-to-end grasp planning for the bin-picking. Part (b) highlights the grasp evaluation process and depicts it for a particular grasp pose instance. Details of the instance segmentation module are provided in Section~\ref{sec:cnn_design}. The grasp pose sampling and grasp pose evaluation are discussed in Section~\ref{sec:method:grasp-planning}.}
    \label{fig:md}
\end{figure*}


\section{Our Method: Bin picking with unknown objects} \label{sec:method}
Our bin-picking framework mainly consists of two parts. One is the CNN-based model training for class-agnostic instance segmentation and another is the grasp-pose planning using the predicted segmentation mask. First, we will describe the grasp-pose planning part in detail. Next, we will iterate through the deep learning framework for the class-agnostic instance segmentation. A schematic diagram of our proposed method is depicted in the Figure~\ref{fig:md} for reference.

\subsection{Class-agnostic Instance Segmentation}
\label{sec:cnn_design}
Our class-agnostic instance segmentation method aims to segment the previously unseen objects in the bin in a real-world setting. For this, a deep-learning-based framework is utilized via sim-2-real transfer learning. The CNN model is trained entirely in the simulation and learning is transferred directly to the real world. The crucial step for setting up any deep learning framework is to acquire the appropriate training data, choose a befitting CNN model based on the application requirement, and perform proper training. We will describe the details of these components, next.

\subsubsection{Data Generation}
For generating the ground truth training samples, the PyBullet robotic simulator has been used. A synthetic environment has been created within the simulator that consists of a bin kept over a table and objects are spawned in it randomly. The simulated camera is kept just above the bin facing downwards at a distance of 70 cm from the bin floor. The number of objects in the scene ranges from 1 to 20 for different samples.

The 3D object models are taken from an open source google-scanned-objects~\cite{gso} repository and consist of daily-use objects such as groceries, medicines, and toys. For each scene, the bin and the scene objects are assigned different textures from a pool of available options. For the bin floor, the textures are pooled from 20 different wooden textures downloaded directly from the web.  For objects, the Describable Textures Dataset~\cite{textures-in-wild} repository is used which consists of 5,640 texture images of 47 different categories. The camera orientation is randomized for each scene within a short range such that the objects in the scene remain in the camera view. The light parameters of the simulation are also randomized representing a range of scene illuminations varying from the bright day scenes to the dark dim-light scenes.

\subsubsection{Network design choice and the training}
For the CNN network design choice, we choose the standard high-performing Mask-RCNN network with ResNet-50 as the backbone. Our proposed bin-picking framework makes use of an open-loop motion planner in which the grasp pose is predicted once and then the robot executes it without further feedback from the vision. Thus, real-time vision feedback is not necessary, however, the quality of the grasp pose matters. The grasp planning algorithm as described in the next section, depends solely on the segmentation mask predicted by the CNN network. Thus, for our bin-picking framework, the segmentation accuracy is more important than the inference time. 

For the training, PyTorch~\cite{pytorch} deep learning library is used. The network was trained for 25 epochs with a batch size of 10. The training was carried out using 3 Nvidia-1080Ti GPUs. During the inference, only 1 GPU is used. The training dataset consists of a total of 30,000 samples and a 9:1 ratio is kept for training and validation sets.

\subsection{Grasp Planning Framework}
\label{sec:method:grasp-planning}
Our bin-picking framework takes the cumulative object instance segmentation mask as input and outputs the final grasp pose for the robot action. The framework for obtaining the instance segmentation mask is described in the previous subsection. To describe our grasp-pose planning method, we define the grasp pose as follows:

\begin{equation}
	    G_i = (P_i, \Theta_i, W_i, Q_i)\label{G_r}
\end{equation}

where $P_i$ represents the center point of the grasp pose $G_i$. $\Theta_i$ denotes the angle of the grasp pose. The grasp pose angle is planner, measured along the vertical axis (i.e. z-axis). The horizontal x-axis is assumed to be the reference zero angle. $W_i$ refers to the width of the grasp pose rectangle, and $Q_i$ represents the grasp quality index.

The grasp pose is calculated in image coordinates and converted into the robot's world Cartesian frame. This conversion requires using intrinsic and extrinsic camera parameters obtained through a standard calibration procedure. The depth values used for this purpose are expressed in the camera's reference frame. The camera is positioned above the workspace bin at a fixed distance, facing downwards. The grasp-pose evaluation method consists of many sub-steps that are executed sequentially. The overall flow of the method is summarised in Algorithm~\ref{algo}. The details of the different components of the method are described, next.
\vspace{0.1mm}
\subsubsection*{1. Sampling Candidate Grasp-pose}
The algorithm samples grasp poses using segmentation masks generated by our category-agnostic segmentation method (Section~\ref{sec:cnn_design}).  At each segmentation instance, \textit{D} number of grasp poses are sampled at equally spaced predefined angles (\textit{D}$=$6 in our case). Each of these grasp-poses $G_i$ is represented by a rectangle of width $gw$ and breadth $gb$ in the image plane. The centers of the segmentation instances become the centers of the corresponding grasp poses. 
 For further processing, the rectangular region corresponding to the grasp pose is cropped from the segmentation mask and horizontally aligned. Then, it is translated such that the top left corner of it coincides with the origin.

\subsubsection*{2. Grasp Pose Subsectors Identification}

To ensure a comprehensive evaluation of a grasp pose $G_i$, our objective is to partition the complete area within the grasp pose rectangle into three distinct subsectors. The tactile contact sector $S_{tc}$ denotes the section of the target object's area within the grasp pose rectangle.  The unobstructed space sector $S_{uo}$ encompasses the region within the grasp pose boundary where the gripping device is unlikely to encounter obstacles during the grasp attempt (i.e. the area corresponding to the background region as per the segmentation mask). The remaining segment constitutes the collision sector $S_{cl}$ indicating the area where the gripping device is prone to collide with other objects. The derivation of these subsectors is done through a simple strategy that uses the obtained segmentation mask. In the grasp pose rectangle area, the pixels corresponding to the target objects are assigned to $S_{tc}$, the pixels corresponding to the background class are assigned to $S_{uo}$ and the rest pixels are assigned to $S_{cl}$. For visualization in Figure~\ref{fig:md2}, the subsectors, namely, $S_{tc}$, $S_{uo}$ and  $S_{cl}$ are depicted with green, white, and red colors. 


\SetKwComment{Comment}{/* }{ */}
\begin{algorithm}
\caption{Grasp planning algorithm as used in our bin-picking framework. The algorithm is discussed in detail in Section~\ref{sec:method:grasp-planning}.}
\label{algo}
\SetKwData{VariableA}{$Dmap$}
\SetKwData{VariableB}{$Smap$}
\SetKwData{VariableC}{$P_i$}
\SetKwData{VariableD}{$\Theta_i$}
\SetKwData{VariableE}{$W_i$}
\SetKwData{VariableF}{$Q_i$}
\SetKwData{VariableG}{$G_i$}
\SetKwData{VariableH}{$W_0$}

\BlankLine
\VariableB $\leftarrow$  Predicted instance segmentation map\;
\VariableC $\leftarrow$  Center point of the grasp pose\;
\VariableD $\leftarrow$  Angle of the grasp pose\;
\VariableE $\leftarrow$  Width of the grasp pose rectangle\;
\VariableH $\leftarrow$  Initial fix width of the grasp pose rectangle\;
\VariableF $\leftarrow$  Grasp quality index (GQI)\;
\VariableG $\leftarrow$  Grasp pose = (\VariableC, \VariableD, \VariableE, \VariableF)\;
Let $i$ denote the i\text{-th} grasp pose parameters and the best grasp pose parameters be denoted by $*$
\BlankLine

\KwData{\VariableB}
\KwResult{Best grasp pose $G^*$}
\BlankLine

\SetKwFunction{FindGraspPoses}{FindGraspPoses}
\SetKwFunction{SelectBestGrasp}{SelectBestGrasp}
\SetKwFunction{SelectBestPose}{SelectBestPose}
\SetKwFunction{PoseEvaluation}{PoseEvaluation}

\SetKwProg{Function}{Function}{:}{end}
\BlankLine
    Initialize $G^*$ = $\emptyset$\;
   
    \ForEach{Mask in \VariableB}{
        $P_i = MaskCentre$\;
        $\Theta_i = MaskAngle$\;
        $candidateGraspPoses$ = append.($P_i$,$\Theta_i$,$W_0$,$NONE$)
        }
    \ForEach{Pose in $candidateGraspPoses$}{
            $G$ = \PoseEvaluation{Pose}\;
            $refinedPoses$ = append.($G$)\;
        }
    $G^*$ = \SelectBestPose($refinedPoses$)\;
    \Return{$G^*$}\;

\BlankLine
\BlankLine

\Function{\PoseEvaluation{Pose}}{
    \BlankLine
    SubsectorsIdentification : $S_{tc}$, $S_{uo}$, $S_{cl}$\;
    \BlankLine
    ValidPose = False\;
    
    ObjectWidth = MaxWidth($S_{tc}$)\;
    FslWidth = minWidth($S_{uo}Left$)\;
    FsrWidth = minWidth($S_{uo}Right$)\;
    
    \BlankLine
    \If{\text{FslWidth} \(>\) \text{GripperFingerWidth} \quad \text{and} \quad \text{FsrWidth} \(>\) \text{GripperFingerWidth} \quad \text{and} \quad \text{ObjectWidth} \(<\)  \text{MaxGripperOpening}}
    {
        ValidPose = True\;
    }
    \Else{Invalid Pose}

    \If{ValidPose}{
        $Pose$.\VariableC = Centre($S_{tc}$)\;
        $Pose$.\VariableE = Centre($S_{uo}Right$) - Centre($S_{uo}Left$)\;
        $Pose$.\VariableF = OSS + CTS + SS\;
        
        \Return{$Pose$}\;
    }

    \Else{\Return{None}}
    
}


\end{algorithm}

\subsubsection*{3. Grasp Pose Filtration}
In the process of validating all the sampled grasp poses, we assess the suitability of each one. A grasp pose is deemed unsuitable under the following circumstances:
For all the sampled grasp poses, we check for the validity of each one. A grasp pose is not suitable in the following two situations: 
\begin{itemize}
    \item If the width of the target object along the orientation of the grasp pose exceeds the maximum potential opening of the gripper, then the grasp becomes unviable. To ascertain this, we compare the maximum width of the tactile contact sector $S_{tc}$ with the gripper's maximum opening capacity.
    \item Adequate space within the free-space sector must be available for the gripper's fingers to enter. To confirm this, we compare the minimum width of the unobstructed space sector $S_{uo}$ on both sides of the grasp pose with the width of the gripper's fingers.
\end{itemize}

\subsubsection*{4. Grasp Pose Finetuning}
We finetune the grasp poses to enhance their effectiveness. Initially, we reposition the grasp pose's center to align with the center of the $S_{tc}$. This adjustment ensures that the grasp pose's central point aligns more accurately with the center of the target object along the grasp pose orientation, resulting in improved stability during the grasping process.
Furthermore, in the process of determining the refined width, we calculate the disparity between the centers of the masks representing the left and right subparts of the region $S_{uo}$. This method allows us to derive a more precise measurement of the refined width, facilitating a more accurate assessment of the available space for the gripper's fingers.

\subsubsection*{5. Grasp Quality Assesment}  If more than one grasp pose
has passed the pose validation step, the grasp quality index $Q_i$ is calculated for each of the valid grasp poses for their
ranking. For the calculation of $Q_i$, three things are taken into
consideration: first is unobstructed-space-score (\texttt{OSS}), which is the normalized area of the unobstructed space sector $S_{uo}$, second is contact-tangibility-score (\texttt{CTS}), which is the normalized area of the tactile contact sector $S_{tc}$ within a predefined rectangular region around the center, and the third is the segmentation score (\texttt{SS}) which is the confidence score predicted by our instance segmentation network. Each of these components takes values between 0 and 100. The value of the grasp quality index $Q_i$ is obtained by taking an average of the above three components.

\begin{table}[t]
    \begin{center}
	\caption{Comparison of two datasets considered for the evaluation of the methods. The two datasets consist of diverse daily-use objects kept within a bin in clutter. WISDOM~\cite{danielczuk2019segmenting} is a public benchmark while Ours is our custom-created dataset. Each dataset provides RGB images, depth maps, and ground truth segmentation of the object instances.}
	\label{table:dataset_compr}
	\begin{tabular}{|l|l|l|} 
	    \hline
            & WISDOM & Ours \\
            \hline
            Depth Sensor & Industrial Grade & Commodity Level \\
            Depth Map noise & Almost noise free & Considerable noise \\
            Depth Accuracy & 25 - 500 micrometers & 2.5 - 5 millimeters \\
            Samples & 300 & 100 \\
            Image Size & 1032x772 & 640x480 \\
            Object Types & Daily use objects & Daily use objects \\
            Segmentation labels & Yes & Yes \\
            \hline
		\end{tabular}
	\end{center}
\end{table}

\section{Results and Discussion}
In this section, we assess the effectiveness of our method for bin-picking unknown novel objects through class-agnostic segmentation. First, in the next subsection, we evaluate our proposed framework for class-agnostic segmentation. Subsequently, an evaluation of the proposed bin-picking method is carried out with real-world bin-picking experiments.

\subsection{Evaluation of the Class-agnostic instance segmentation method}
\label{sec:cas_results}
To evaluate our proposed framework for class-agnostic instance segmentation for bin-picking applications, two datasets are considered. First, is \textit{WISDOM}~\cite{danielczuk2019segmenting} which is a public benchmark dataset in this domain, and second is our custom-made dataset.
In Table~\ref{table:dataset_compr} the two datasets are compared over various attributes. The notable difference between the two datasets is the depth sensors used for capturing the depth maps. While for the WISDOM dataset, industrial grade costlier Phoxi camera is used that produces high precision (accuracy of 25-500 um) depth maps, for our custom-dataset the commodity level cost-effective Realsense camera is used that produces considerable noisy depth maps (accuracy of 2.5 - 5 mm). 

As an evaluation metric, the average precision (AP) and average recall (AR) are used as defined by the COCO benchmark~\cite{cocodataset} for the instance segmentation task. For calculating AP and AR, IoU thresholds from 0.50 to 0.95 with a 0.05 margin were used and top-100 detections were considered. 
The experimental results are reported in Table~\ref{table:compare}. Our method has achieved better results compared to the considered baseline methods~\cite{danielczuk2019segmenting, cas_close_baseline, baseline_icra22}. For baselines, we only consider class-agnostic segmentation works that are related to the bin-picking applications. All the baselines use their respective custom-generated simulated data for the training. The baseline~\cite{danielczuk2019segmenting} uses only a depth map as the input. The baseline~\cite{cas_close_baseline} uses a fusion of depth and RGB features as the input. For~\cite{baseline_icra22}, two variants are considered, one uses only depth map as the input and another uses photo-realistic rendered RGB images.

As shown in the table, the method that has used depth data as the input performs considerably over the Wisdom dataset while the performance over our custom dataset is poorer. The noisy depth maps are the reason behind the performance decline of these methods. Our method used only synthetic RGB images as the input and was able to transfer well in the real world with the help of domain randomization. Our method achieves state-of-the-art performance over both datasets while using only the RGB image as the input. Photo-realism can also be an alternative for smooth sim-to-real transfer as shown by the results of the method~\cite{baseline_icra22}. Nevertheless, this approach mandates meticulously crafted simulations and significant computational expenses, resulting in limited adaptability and impracticality for real-world implementations.
\begin{table}[t]
    \begin{center}
	\caption{Comparing different model variants based on the segmentation performance evaluated in terms of average precision and average recall  (as defined by  COCO benchmarks~\cite{cocodataset}). The performance was measured over the two datasets, namely WISDOM~\cite{danielczuk2019segmenting} and Ours (See Table~\ref{table:dataset_compr}). Our method outperforms all the considered state-of-the-art baselines. For further discussion, please refer to Section~\ref{sec:cas_results}.}
	\label{table:compare}
	\begin{tabular}{l|l|l|c|c|c|c} 
	    \hline
		 Author & Sim-2-real  & Input & \multicolumn{2}{c|}{WISDOM}  & \multicolumn{2}{c}{Ours} \\
		 & method & &  \multicolumn{1}{c}{AP} & \multicolumn{1}{c|}{AR} & \multicolumn{1}{c}{AP} & \multicolumn{1}{c}{AR}  \\
		\hline
        

        \cite{danielczuk2019segmenting} & Direct & D &  51.6 & 64.7 & 0.009 & 0.085\\
        \hline
        \cite{cas_close_baseline} & Noise-Modeling & D+RGB & 60.5 & 66.4 &  65.5 & 70.2 \\
         \hline
        \cite{baseline_icra22} & Direct & D & 56.5 & 65.4 & 0.011 & 0.098 \\
         & Photo-realism & RGB & 60.6 & 68.8 & 67.8 & 73.6 \\
        \hline
        & Direct & D & 59.8 & 67.5 & 0.013 & 0.103 \\
         & Direct & RGB & 41.9 & 56.3 & 55.1 & 60.4 \\
        \textbf{Ours} & Direct & L & 40.8 & 56.3 & 52.5 & 57.3 \\ 
        & DR & L & 61.2 & 69.1 & 66.3 & 72.4 \\
         & \textbf{DR} & \textbf{RGB} & \textbf{62.1} & \textbf{69.2} & \textbf{72.8} & \textbf{77.3} \\
        
        \hline


		\end{tabular}
	\end{center}
\end{table}

\begin{table}[h!]
    \begin{center}
	\caption{Real-world bin-picking experiments with opaque and non-opaque objects in a challenging scenario. Our method outperforms the considered state-of-the-art baselines in each category. The performance is measured in terms of grasp success rate. For the grasp trials, a maximum of 15 and 10 objects are present in the bin, respectively for opaque and non-opaque categories. For further details, please refer to Section~\ref{sec:bin-picking-exps}. (D stands for depth in the table.)}
	\label{table:bin-picking}
	\begin{tabular}{l|l|c|c|c} 
	    \hline
		 Method & Input & \multicolumn{2}{c|}{Grasp Success}  & Inference \\
        & Type & \multicolumn{1}{c}{Opaque} & \multicolumn{1}{c|}{Non-opaque} &  Time (ms) \\
		\hline

       GRConv~\cite{antipodal} & RGB+D & 41 & 34 & 21 \\
       DexNet (4.0-PJ)~\cite{dexnet4} & D & 66 & 54 & 1212 \\
       Raj~\cite{case_baseline} & D & 87 & 73 & 375 \\
       \hline
       \textbf{Ours} & \textbf{RGB} &  \textbf{98} & \textbf{97} & \textbf{144 }\\
        
        \hline
		\end{tabular}
	\end{center}
\end{table}

\begin{table*}[t]
    \centering
    \caption{Qualitative examples of instance-segmentation and grasp pose prediction by our method (i.e. rows 4 and 8, respectively) along with the grasp pose predictions by the considered baseline methods (i.e. rows 5-7). Scenes 1-3 consist of \textbf{opaque} objects and scenes 4-6 consist of \textbf{non-opaque} (i.e. transparent and translucent objects). The results indicate that our grasp-pose planning framework demonstrates superior performance in predicting high-quality grasp-poses compared to existing approaches, particularly in difficult bin-picking scenarios (Best viewed in color).}
    \begin{tabular}{M{10mm}|M{19mm}M{19mm}M{19mm}M{19mm}M{19mm}M{19mm}M{19mm}}
       \toprule
          \textbf{Scenes} & \textbf{RGB} & \textbf{Depth visualization} & \textbf{Segmentation (Ours)} & \textbf{GR-Conv}~\cite{antipodal} & \textbf{DexNet}~\cite{dexnet4} & \textbf{Raj et. al.}~\cite{case_baseline} &  \textbf{Ours} \\
        \midrule
        \textbf{Scene-1}  &
        \includegraphics[scale=0.1]{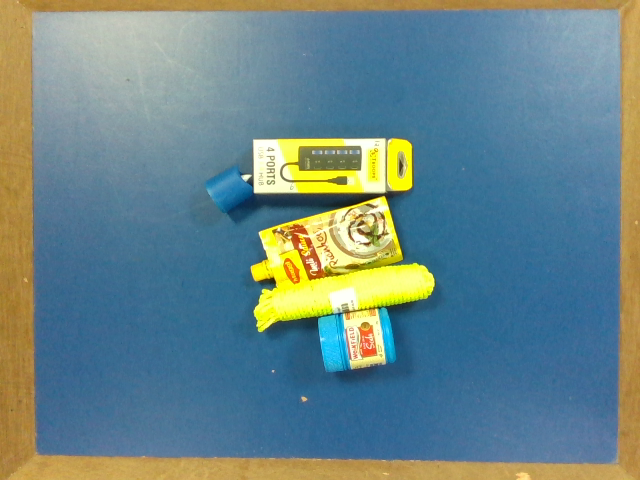} &
        \includegraphics[scale=0.1]{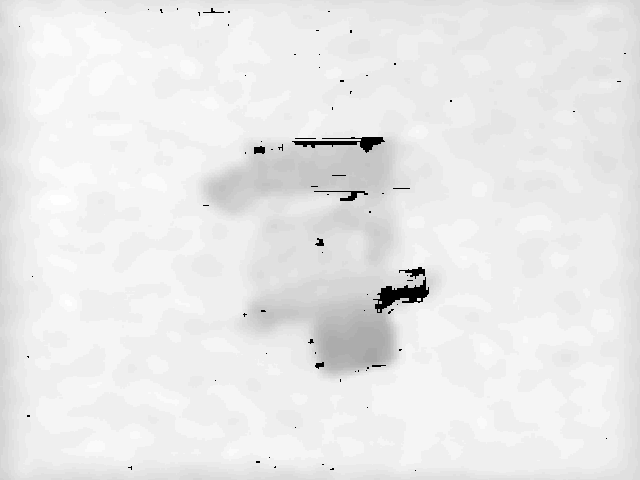} &
        \includegraphics[scale=0.1]{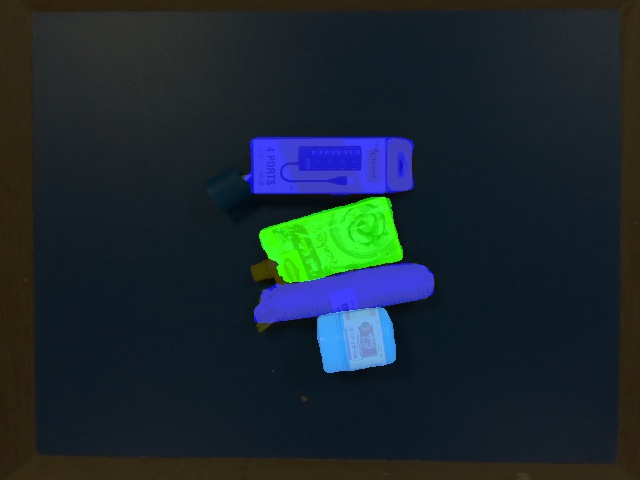} &
        \includegraphics[scale=0.1]{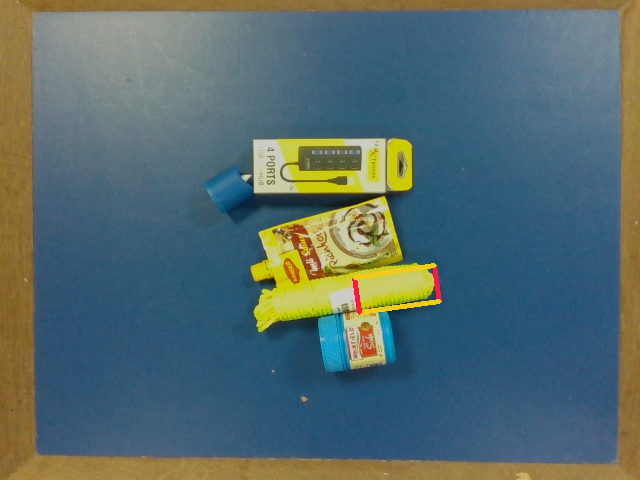} &
        \includegraphics[scale=0.1]{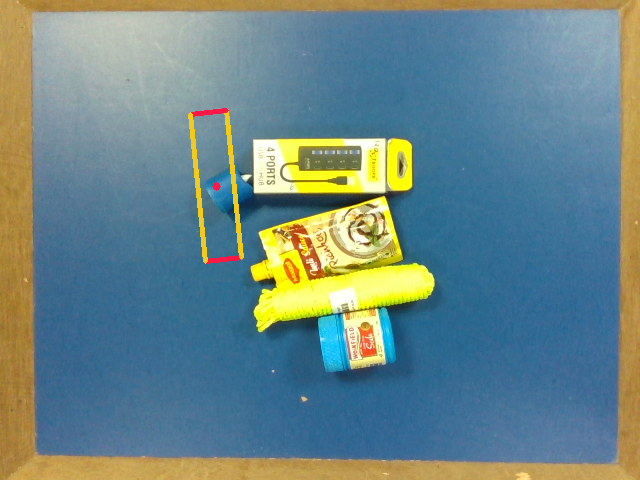} & \includegraphics[scale=0.1]{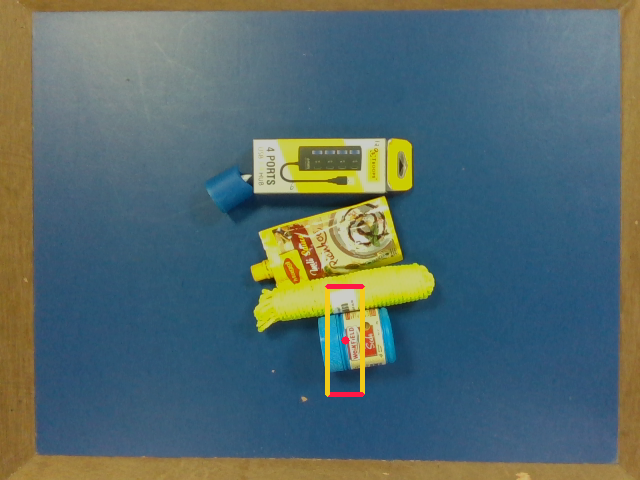} &
         \includegraphics[scale=0.1]{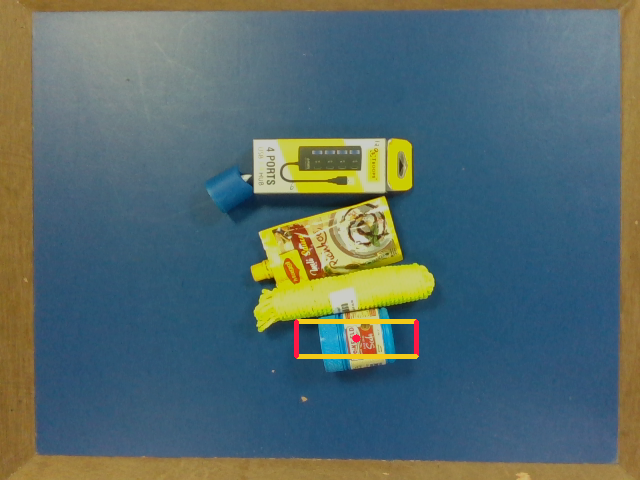} \\ \\

        \textbf{Scene-2}  &
        \includegraphics[scale=0.1]{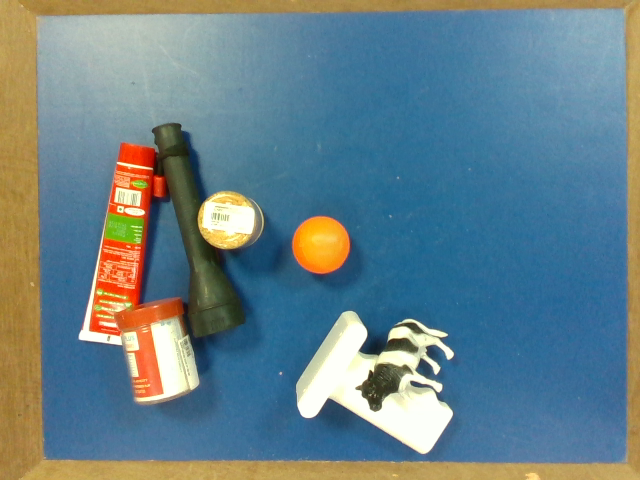} &
        \includegraphics[scale=0.1]{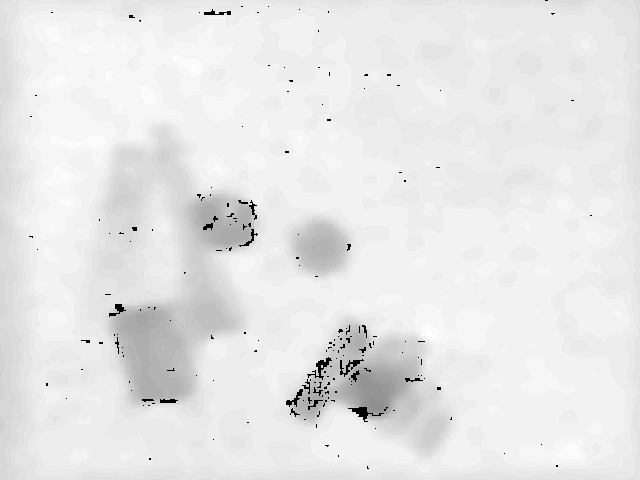} &
        \includegraphics[scale=0.1]{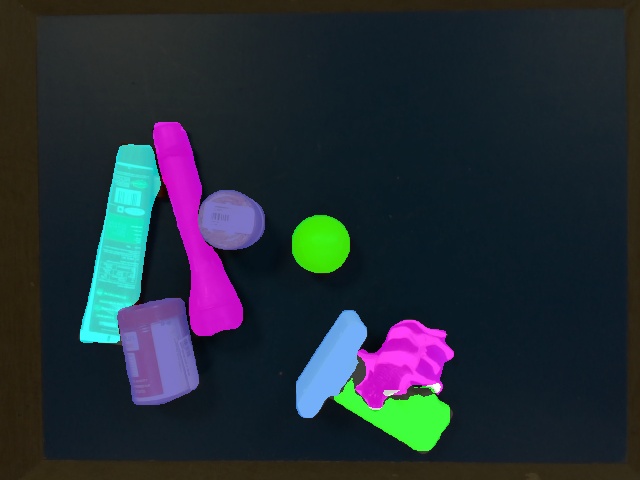} &
        \includegraphics[scale=0.1]{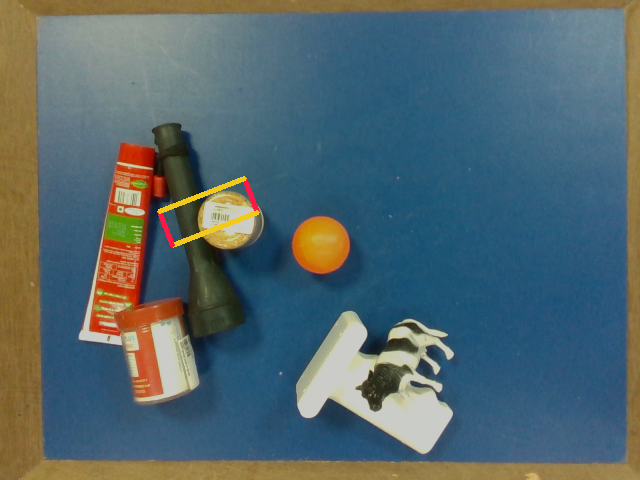} &
        \includegraphics[scale=0.1]{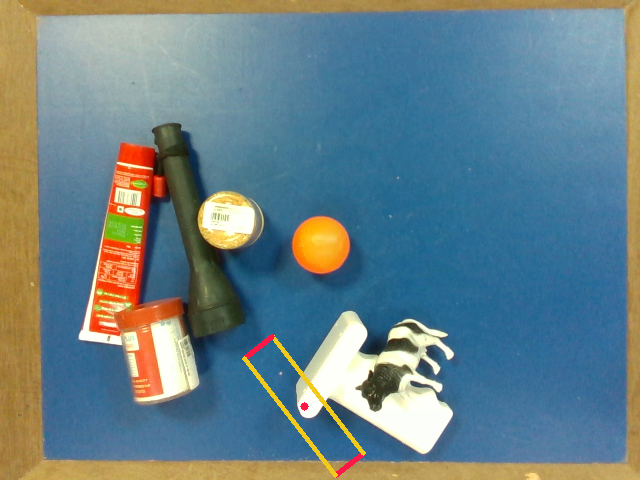} & \includegraphics[scale=0.1]{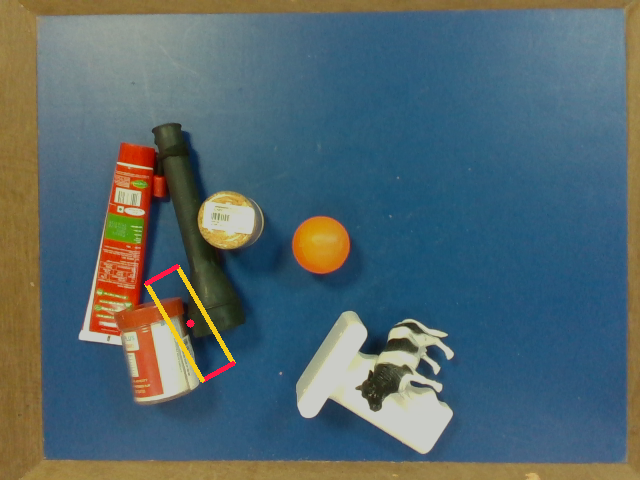} &
         \includegraphics[scale=0.1]{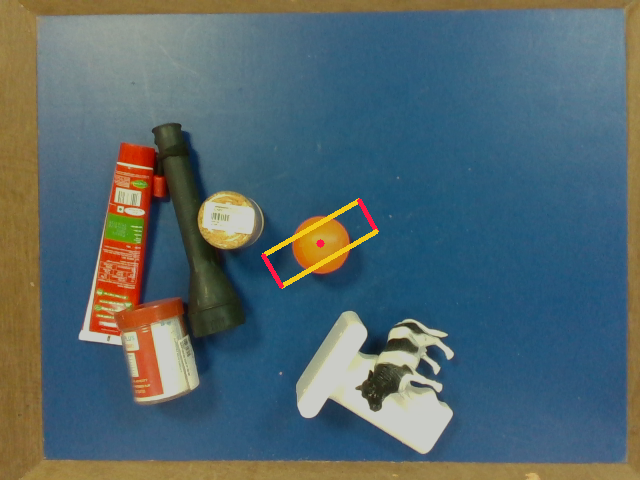} \\ \\

         \textbf{Scene-3}  &
        \includegraphics[scale=0.1]{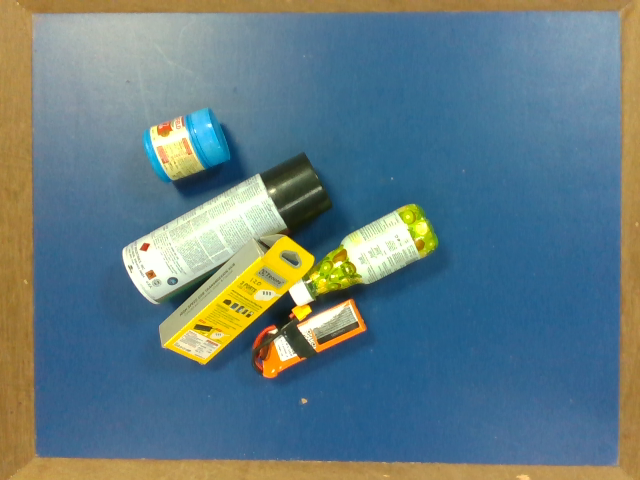} &
        \includegraphics[scale=0.1]{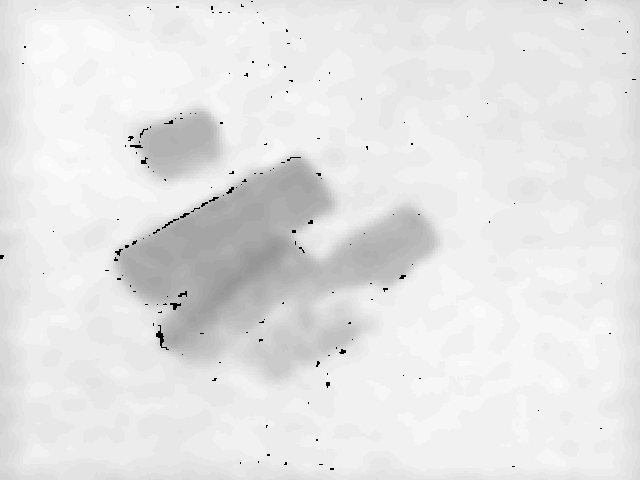} &
        \includegraphics[scale=0.1]{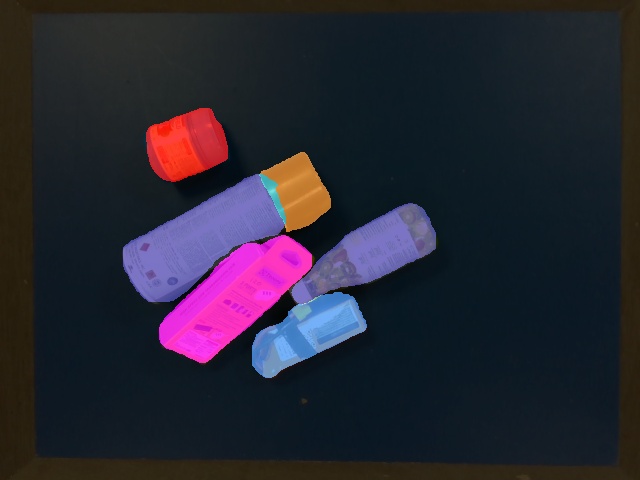} &
        \includegraphics[scale=0.1]{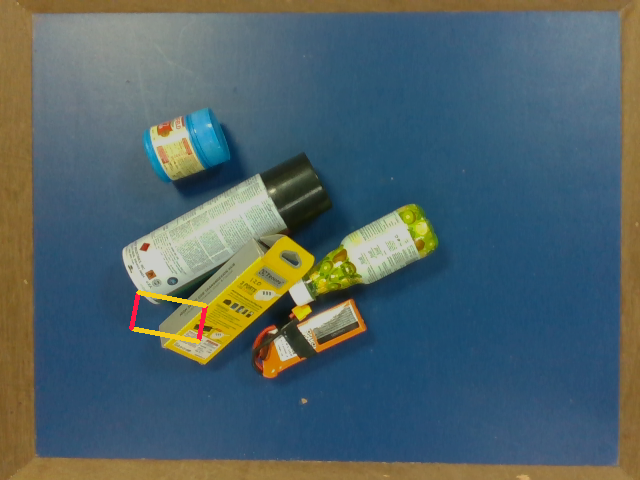} &
        \includegraphics[scale=0.1]{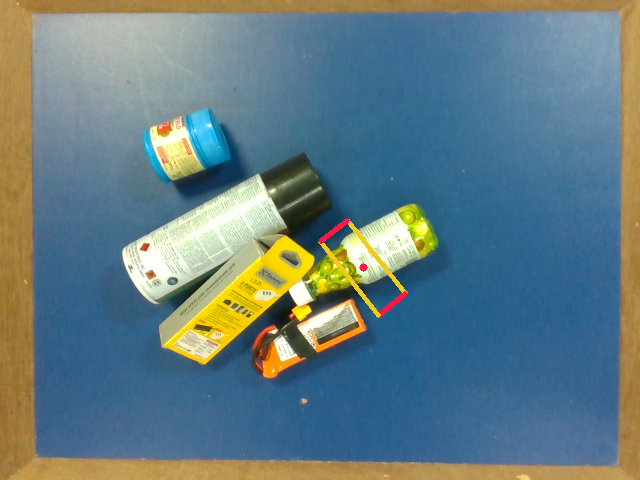} & \includegraphics[scale=0.1]{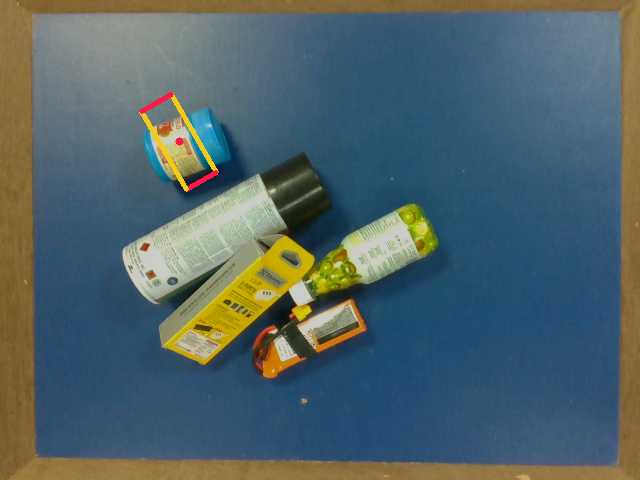} &
         \includegraphics[scale=0.1]{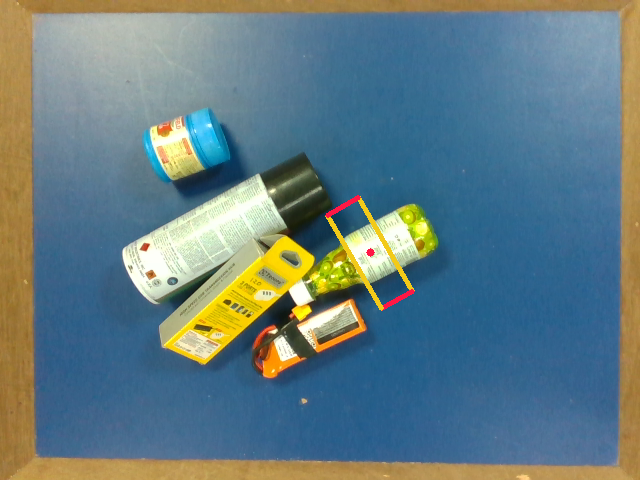} \\ \\

          \textbf{Scene-4}  &
        \includegraphics[scale=0.1]{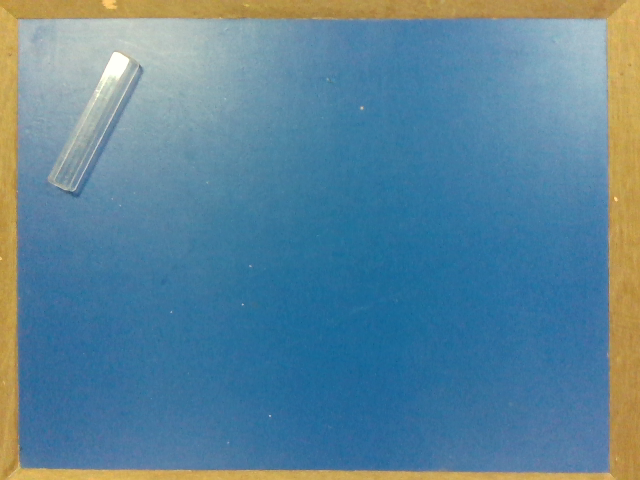} &
        \includegraphics[scale=0.1]{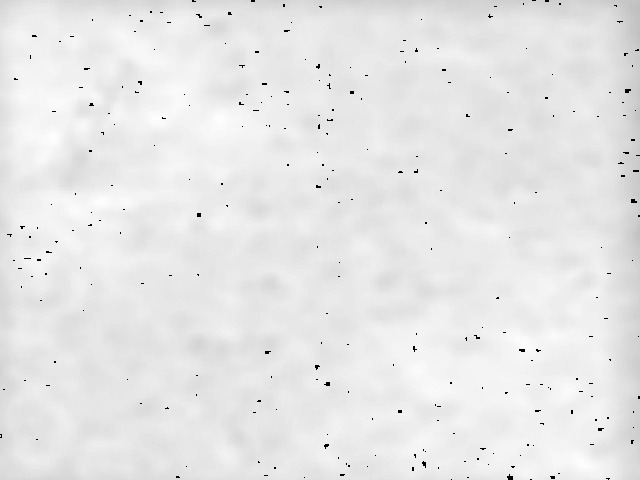} &
        \includegraphics[scale=0.1]{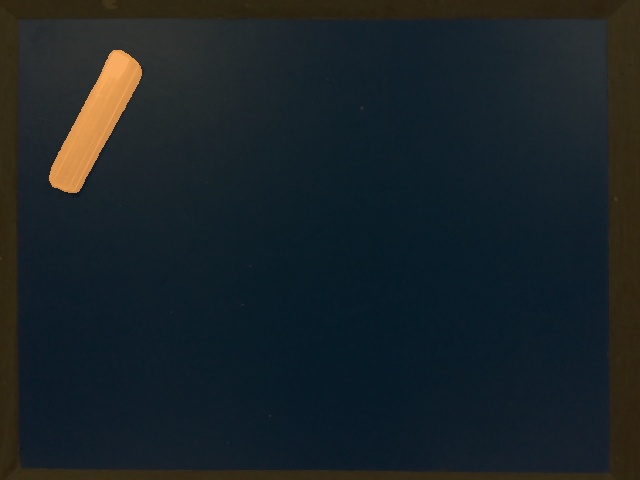} &
        \includegraphics[scale=0.1]{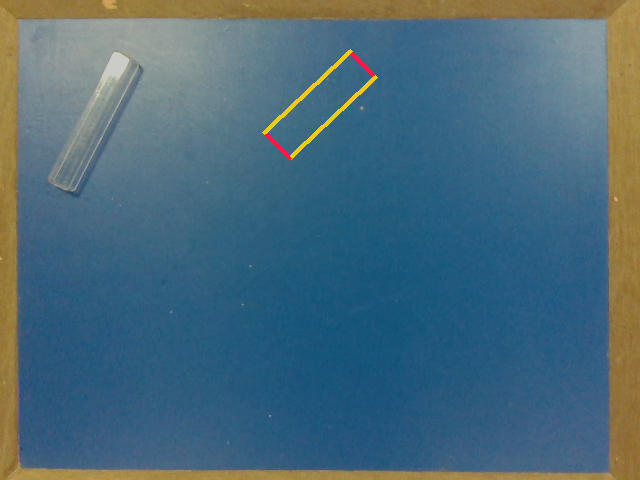} &
        \includegraphics[scale=0.1]{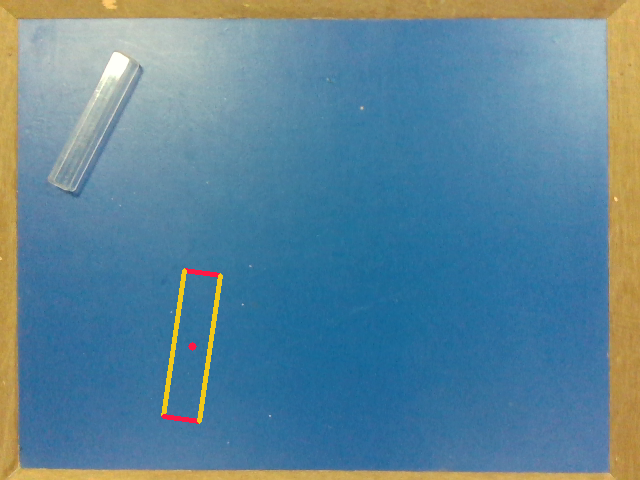} & \includegraphics[scale=0.1]{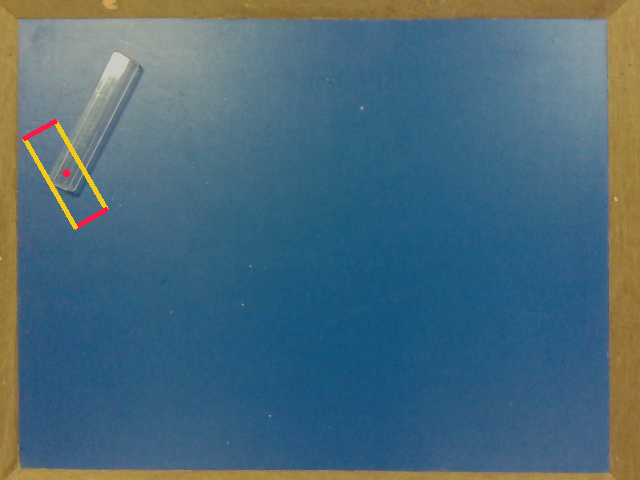} &
         \includegraphics[scale=0.1]{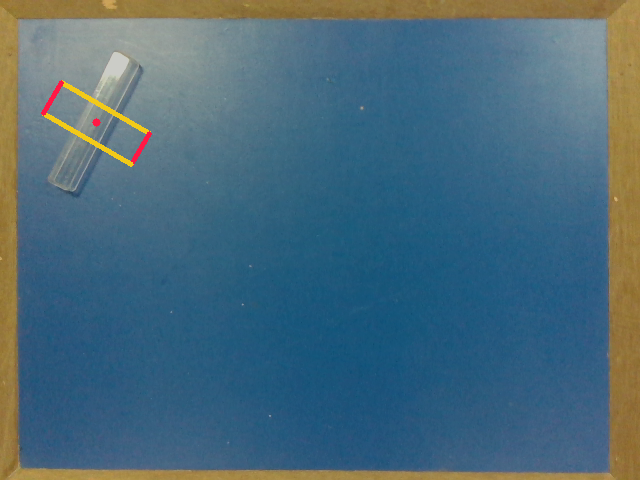} \\ \\

          \textbf{Scene-5}  &
        \includegraphics[scale=0.1]{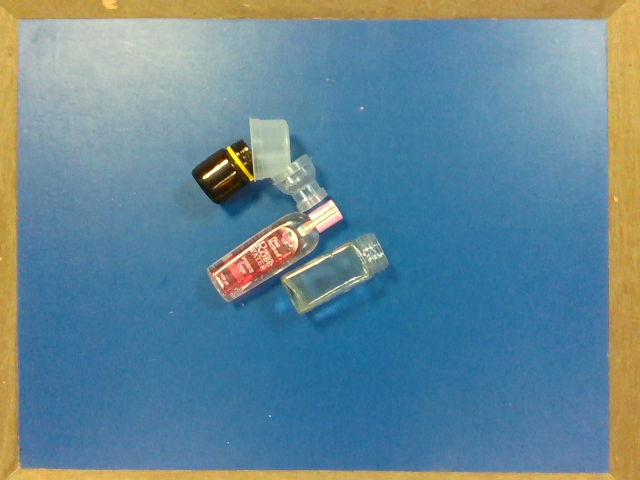} &
        \includegraphics[scale=0.1]{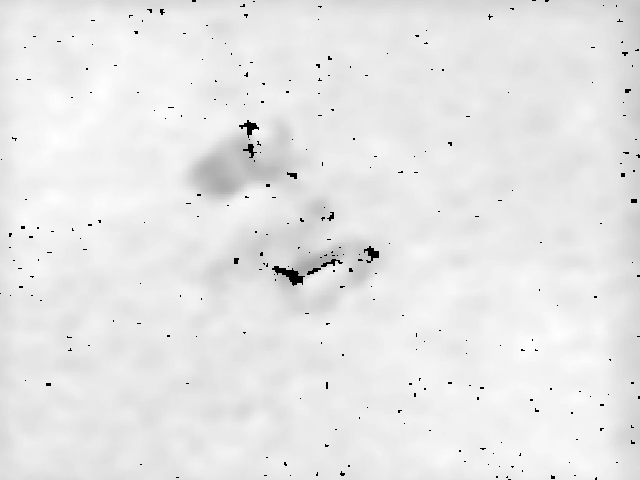} &
        \includegraphics[scale=0.1]{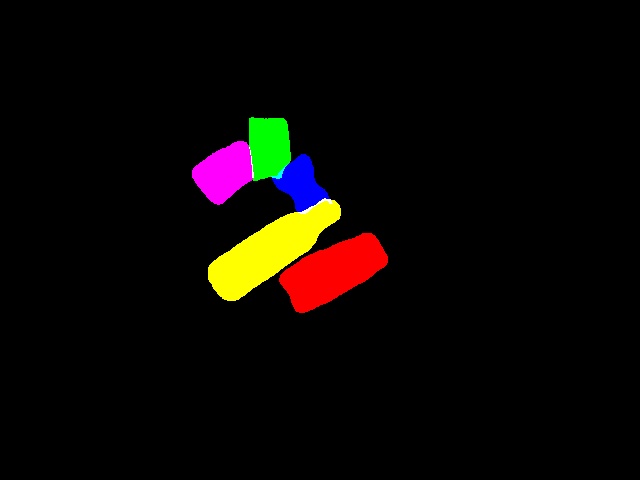} &
        \includegraphics[scale=0.1]{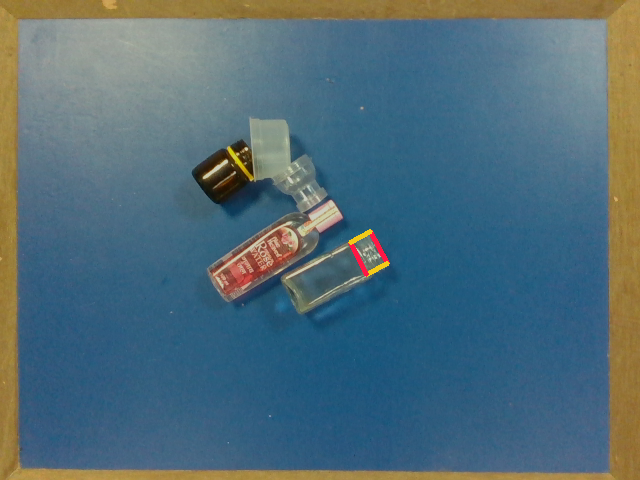} &
        \includegraphics[scale=0.1]{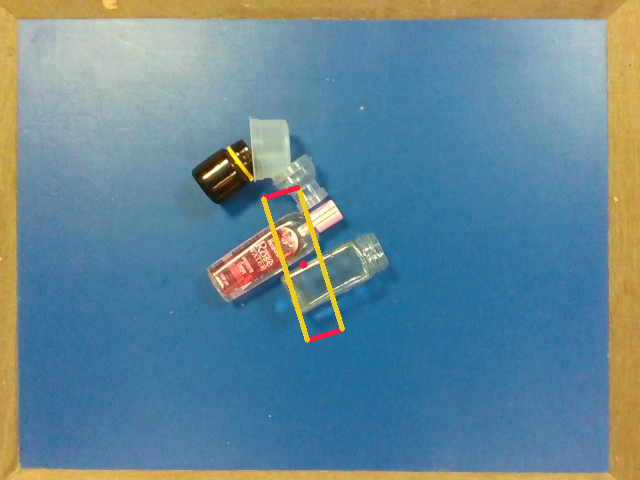} & \includegraphics[scale=0.1]{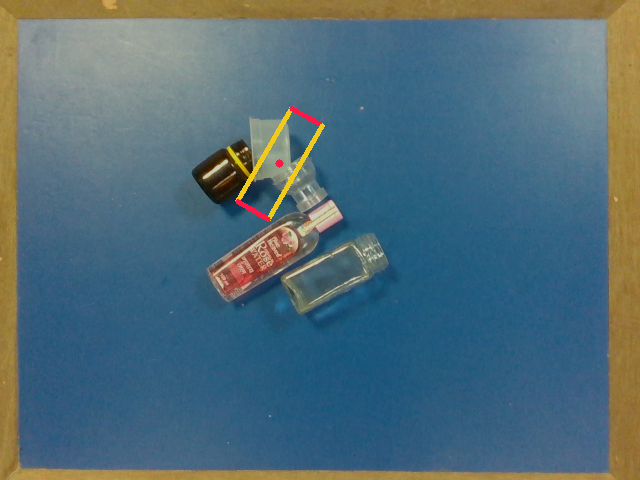} &
         \includegraphics[scale=0.1]{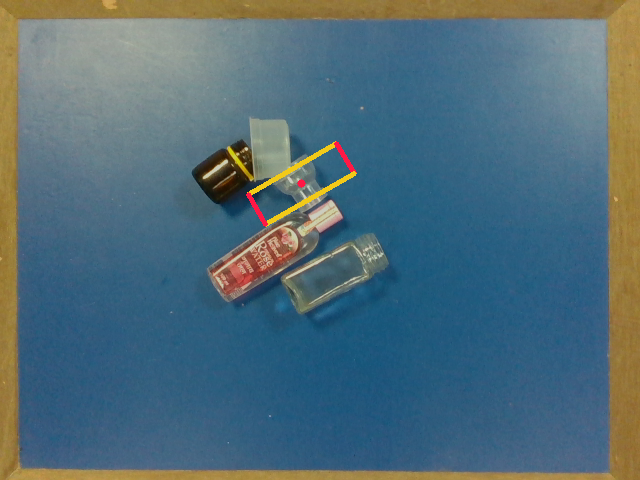} \\ \\

          \textbf{Scene-6}  &
        \includegraphics[scale=0.1]{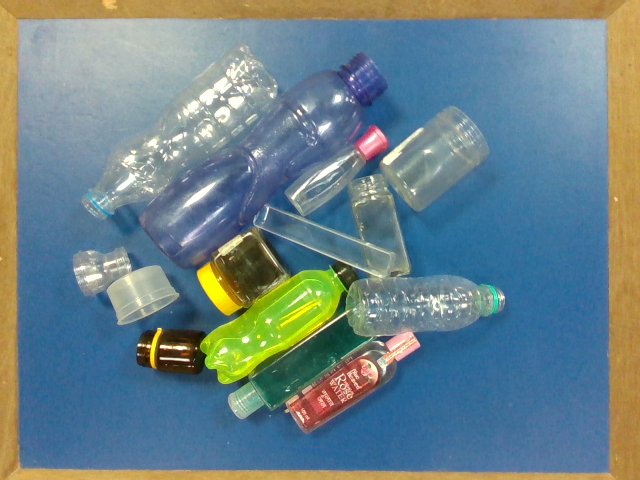} &
        \includegraphics[scale=0.1]{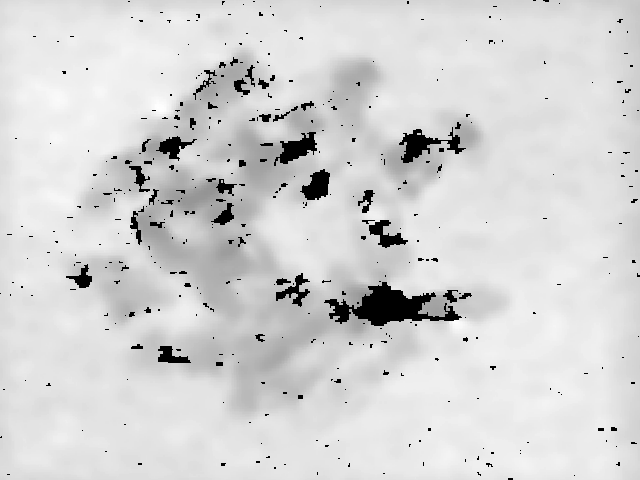} &
        \includegraphics[scale=0.1]{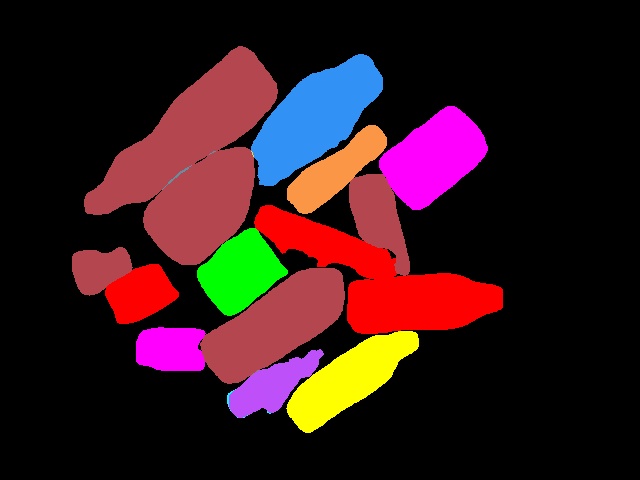} &
        \includegraphics[scale=0.1]{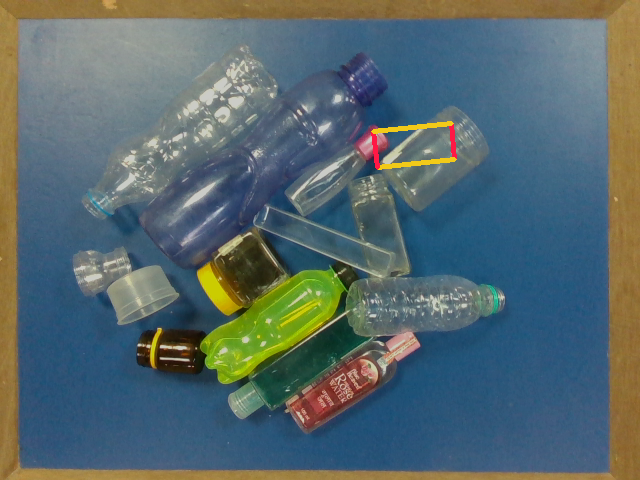} &
        \includegraphics[scale=0.1]{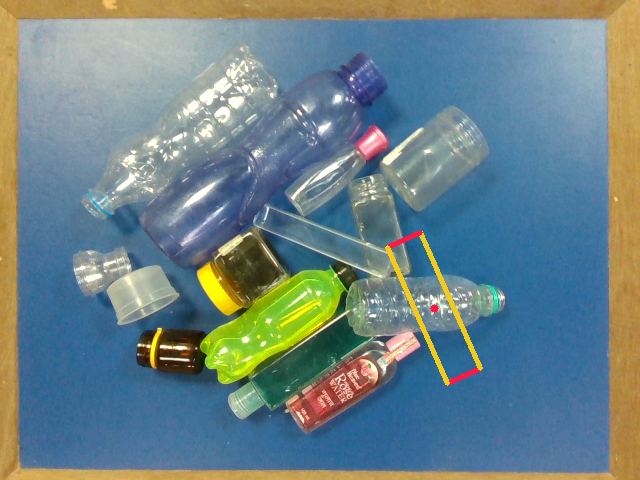} & \includegraphics[scale=0.1]{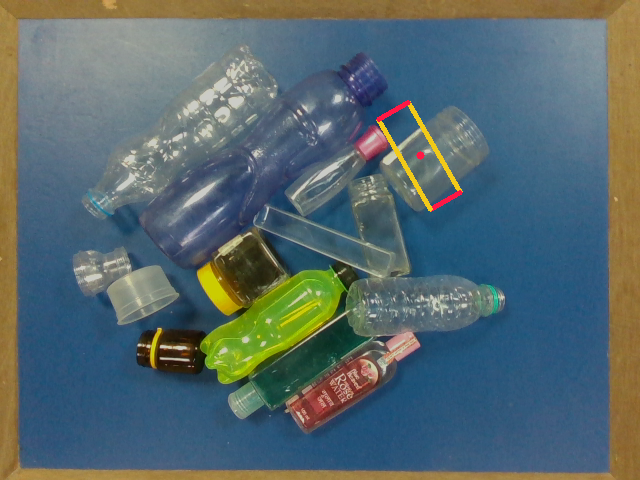} &
         \includegraphics[scale=0.1]{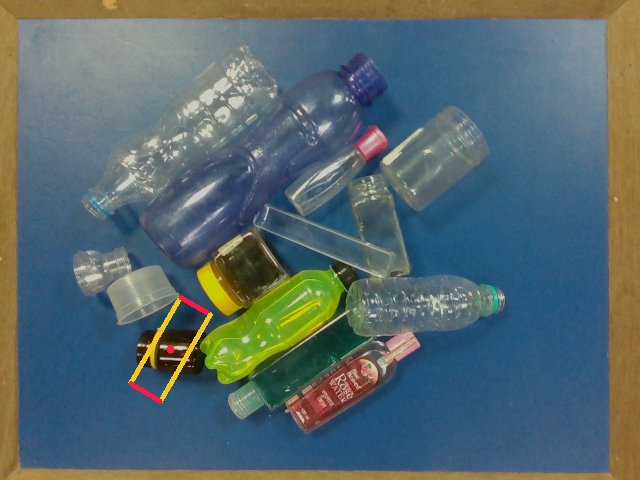} \\ \\

        \bottomrule
    \end{tabular}
    \label{tab:qualitative-overall}
    \end{table*}


\subsection{Bin Picking Experiments}
\label{sec:bin-picking-exps}
To evaluate the complete end-to-end bin-picking pipeline, we have performed real-world experiments. For the experiments, a UR5 robotic manipulator arm is used. A Realsense D435i RGB-D camera is mounted upon the wrist of the manipulator as an eye-in-hand configuration. As a gripping tool, the Schunk WSG-50 gripper (two-fingered parallel-jaw gripper) is mounted at the end-effector of the manipulator. The setup is shown in Figure~\ref{fig:intro}. As shown in the figure, the camera is looking directly in the downward direction where the target objects are placed in a bin.

The experiments consist of various daily usable objects including transparent and semi-transparent objects. Our grasp prediction method mainly relies on the segmentation mask generated from the CNN network, which uses only the RGB image. All the grasp-pose parameters are calculated without using the depth map. 

For the experiments, we divide the object set into two. One is opaque objects and the other is non-opaque objects (transparent and translucent objects). We evaluate our method along with the considered baseline methods~\cite{antipodal, dexnet4, case_baseline} in two different scenarios, all opaque objects only and all non-opaque objects only. 

The bin-picking experiments are performed using our proposed method and the selected state-of-the-art baseline methods~\cite{case_baseline, dexnet4, antipodal}. For each method, a total of 100 grasp trials are performed in each scenario type (i.e. opaque and non-opaque). Initially, 15 and 10 objects are randomly thrown in the bin, respectively for opaque and non-opaque categories. Then, objects are grasped one by one and put into the receptacle. Again, a new iteration is started when either two consecutive failures have occurred or all the objects in the scene are grasped. Consecutive grasp failures at the same location are counted only once. Thus, this process is repeated until the total number of grasp attempts reaches 100.

The results are reported in Table~\ref{table:bin-picking}. Our method outperforms all the considered baselines by a large margin. The methods~\cite{antipodal} and \cite{dexnet4} were trained over noise-free simulated depth images and thus perform poorly in noisy depth sensing environments. In the case of non-opaque objects, the noise in the depth maps increases further resulting in further decline in the performance. Our method is independent of the depth data and thus performs superior. Furthermore, it is interesting to see that our method performs equally well with non-opaque objects although the training data for our instance-segmentation network does not contain any non-opaque objects.

\section{Conclusion and Future Works}
This study addresses the critical challenge of category-agnostic instance segmentation for robotic manipulation, enabling versatile applications such as bin-picking with unknown objects in clutter. By focusing on object-centric segmentation and leveraging simulation-based training, our approach is able to segment unknown objects in the real world without a single real-world training sample. The devised strategy effectively addresses the inherent noise in depth sensors and enables reliable picking of objects in the absence of high-precision depth sensing. Notably, our solution accommodates transparent and semi-transparent objects, historically challenging for depth-based techniques. The contributions encompass a successful domain randomization strategy, the provision of benchmark datasets for warehouse applications, and an integrated bin-picking framework for enhanced efficiency.

One of the challenges our method faces is that the segmentation quality becomes poorer when the clutter in the bin rises beyond a certain level. It will be interesting to see a method that can selectively segment the objects reliably that are graspable and avoids others that are mostly occluded and will not likely cause any collision during the grasp attempt. Another possible direction for future work is to encompass the depth of information within the learning process for the instance segmentation task while the input to the deep network is still the RGB image only. One way to achieve this is to add a depth estimation as the auxiliary task in the network design.


 

\bibliographystyle{IEEEtran}
\bibliography{IEEEabrv,refs.bib}

\begin{thebibliography}{10}
\providecommand{\url}[1]{#1}
\csname url@samestyle\endcsname
\providecommand{\newblock}{\relax}
\providecommand{\bibinfo}[2]{#2}
\providecommand{\BIBentrySTDinterwordspacing}{\spaceskip=0pt\relax}
\providecommand{\BIBentryALTinterwordstretchfactor}{4}
\providecommand{\BIBentryALTinterwordspacing}{\spaceskip=\fontdimen2\font plus
\BIBentryALTinterwordstretchfactor\fontdimen3\font minus \fontdimen4\font\relax}
\providecommand{\BIBforeignlanguage}[2]{{%
\expandafter\ifx\csname l@#1\endcsname\relax
\typeout{** WARNING: IEEEtran.bst: No hyphenation pattern has been}%
\typeout{** loaded for the language `#1'. Using the pattern for}%
\typeout{** the default language instead.}%
\else
\language=\csname l@#1\endcsname
\fi
#2}}
\providecommand{\BIBdecl}{\relax}
\BIBdecl

\bibitem{danielczuk2019segmenting}
M.~Danielczuk, M.~Matl, S.~Gupta, A.~Li, A.~Lee, J.~Mahler, and K.~Goldberg, ``Segmenting unknown 3d objects from real depth images using mask r-cnn trained on synthetic data,'' in \emph{2019 International Conference on Robotics and Automation (ICRA)}.\hskip 1em plus 0.5em minus 0.4em\relax IEEE, 2019, pp. 7283--7290.

\bibitem{segmentation_traditional1}
D.~Morrison, A.~W. Tow, M.~Mctaggart, R.~Smith, N.~Kelly-Boxall, S.~Wade-Mccue, J.~Erskine, R.~Grinover, A.~Gurman, T.~Hunn \emph{et~al.}, ``Cartman: The low-cost cartesian manipulator that won the amazon robotics challenge,'' in \emph{2018 IEEE International Conference on Robotics and Automation (ICRA)}.\hskip 1em plus 0.5em minus 0.4em\relax IEEE, 2018, pp. 7757--7764.

\bibitem{segmentation_traditional2}
A.~Kumar and L.~Behera, ``Semi supervised deep quick instance detection and segmentation,'' in \emph{2019 International Conference on Robotics and Automation (ICRA)}.\hskip 1em plus 0.5em minus 0.4em\relax IEEE, 2019, pp. 8325--8331.

\bibitem{dexnet2}
J.~Mahler, J.~Liang, S.~Niyaz, M.~Laskey, R.~Doan, X.~Liu, J.~A. Ojea, and K.~Goldberg, ``Dex-net 2.0: Deep learning to plan robust grasps with synthetic point clouds and analytic grasp metrics,'' in \emph{in Proceedings of Robotics: Science and Systems (RSS)}.\hskip 1em plus 0.5em minus 0.4em\relax IEEE, 2017.

\bibitem{dexnet4}
J.~Mahler, M.~Matl, V.~Satish, M.~Danielczuk, B.~DeRose, S.~McKinley, and K.~Goldberg, ``Learning ambidextrous robot grasping policies,'' \emph{Science Robotics}, vol.~4, no.~26, 2019.

\bibitem{case_baseline}
P.~Raj, A.~Kumar, V.~Sanap, T.~Sandhan, and L.~Behera, ``Towards object agnostic and robust 4-dof table-top grasping,'' in \emph{IEEE 18th International Conference on Automation Science and Engineering (CASE)}, 2022.

\bibitem{graspfusionnet}
W.~Wang, W.~Liu, J.~Hu, Y.~Fang, Q.~Shao, and J.~Qi, ``Graspfusionnet: a two-stage multi-parameter grasp detection network based on rgb--xyz fusion in dense clutter,'' \emph{Machine Vision and Applications}, vol.~31, no.~7, pp. 1--19, 2020.

\bibitem{suction_2022uncertainty}
K.~Tung, J.~Su, J.~Cai, Z.~Wan, and H.~Cheng, ``Uncertainty-based exploring strategy in densely cluttered scenes for vacuum cup grasping,'' in \emph{2022 International Conference on Robotics and Automation (ICRA)}.\hskip 1em plus 0.5em minus 0.4em\relax IEEE, 2022, pp. 3483--3489.

\bibitem{coco}
T.-Y. Lin, M.~Maire, S.~Belongie, J.~Hays, P.~Perona, D.~Ramanan, P.~Doll{\'a}r, and C.~L. Zitnick, ``Microsoft coco: Common objects in context,'' in \emph{Computer Vision--ECCV 2014: 13th European Conference, Zurich, Switzerland, September 6-12, 2014, Proceedings, Part V 13}.\hskip 1em plus 0.5em minus 0.4em\relax Springer, 2014, pp. 740--755.

\bibitem{raj2020learning}
P.~Raj, V.~P. Namboodiri, and L.~Behera, ``Learning to switch cnns with model agnostic meta learning for fine precision visual servoing,'' in \emph{2020 IEEE/RSJ International Conference on Intelligent Robots and Systems (IROS)}.\hskip 1em plus 0.5em minus 0.4em\relax IEEE, 2020, pp. 10\,210--10\,217.

\bibitem{sim2real}
D.~Horv{\'a}th, G.~Erd{\H{o}}s, Z.~Istenes, T.~Horv{\'a}th, and S.~F{\"o}ldi, ``Object detection using sim2real domain randomization for robotic applications,'' \emph{IEEE Transactions on Robotics}, 2022.

\bibitem{baseline_corl20}
C.~Xie, Y.~Xiang, A.~Mousavian, and D.~Fox, ``The best of both modes: Separately leveraging rgb and depth for unseen object instance segmentation,'' in \emph{Conference on robot learning}.\hskip 1em plus 0.5em minus 0.4em\relax PMLR, 2020, pp. 1369--1378.

\bibitem{hofer2021sim2real}
S.~H{\"o}fer, K.~Bekris, A.~Handa, J.~C. Gamboa, M.~Mozifian, F.~Golemo, C.~Atkeson, D.~Fox, K.~Goldberg, J.~Leonard \emph{et~al.}, ``Sim2real in robotics and automation: Applications and challenges,'' \emph{IEEE transactions on automation science and engineering}, vol.~18, no.~2, pp. 398--400, 2021.

\bibitem{baseline_corl21}
Y.~Xiang, C.~Xie, A.~Mousavian, and D.~Fox, ``Learning rgb-d feature embeddings for unseen object instance segmentation,'' in \emph{Conference on Robot Learning}.\hskip 1em plus 0.5em minus 0.4em\relax PMLR, 2021, pp. 461--470.

\bibitem{baseline_tro21}
C.~Xie, Y.~Xiang, A.~Mousavian, and D.~Fox, ``Unseen object instance segmentation for robotic environments,'' \emph{IEEE Transactions on Robotics}, vol.~37, no.~5, pp. 1343--1359, 2021.

\bibitem{baseline_icra22}
S.~Back, J.~Lee, T.~Kim, S.~Noh, R.~Kang, S.~Bak, and K.~Lee, ``Unseen object amodal instance segmentation via hierarchical occlusion modeling,'' in \emph{2022 International Conference on Robotics and Automation (ICRA)}.\hskip 1em plus 0.5em minus 0.4em\relax IEEE, 2022, pp. 5085--5092.

\bibitem{cas_close_baseline}
S.~Back, J.~Kim, R.~Kang, S.~Choi, and K.~Lee, ``Segmenting unseen industrial components in a heavy clutter using rgb-d fusion and synthetic data,'' in \emph{2020 IEEE International Conference on Image Processing (ICIP)}, 2020, pp. 828--832.

\bibitem{domain-randomization}
J.~Tobin, R.~Fong, A.~Ray, J.~Schneider, W.~Zaremba, and P.~Abbeel, ``Domain randomization for transferring deep neural networks from simulation to the real world,'' in \emph{2017 IEEE/RSJ international conference on intelligent robots and systems (IROS)}.\hskip 1em plus 0.5em minus 0.4em\relax IEEE, 2017, pp. 23--30.

\bibitem{he2017mask}
K.~He, G.~Gkioxari, P.~Dollar, and R.~Girshick, ``Mask r-cnn,'' in \emph{Proceedings of the IEEE International Conference on Computer Vision (ICCV)}, 2017.

\bibitem{ren2015faster}
S.~Ren, K.~He, R.~Girshick, and J.~Sun, ``Faster r-cnn: Towards real-time object detection with region proposal networks,'' in \emph{Proceedings of the Conference on Neural Information Processing Systems (NIPS)}, 2015.

\bibitem{zhang2018shufflenet}
X.~Zhang, X.~Zhou, M.~Lin, and J.~Sun, ``Shufflenet v2: Practical guidelines for efficient cnn architecture design,'' in \emph{Proceedings of the European Conference on Computer Vision (ECCV)}, 2018.

\bibitem{fpn}
T.-Y. Lin, P.~Doll{'a}r, R.~Girshick, K.~He, B.~Hariharan, and S.~Belongie, ``Feature pyramid networks for object detection,'' in \emph{Proceedings of the IEEE Conference on Computer Vision and Pattern Recognition (CVPR)}, 2017.

\bibitem{chen2021big}
J.~Chen, S.~Kornblith, M.~Norouzi, and G.~Hinton, ``Big transfer (bit): General visual representation learning,'' in \emph{Proceedings of the International Conference on Learning Representations (ICLR)}, 2021.

\bibitem{panoptic}
A.~Kirillov, R.~Girshick, K.~He, and P.~Doll{'a}r, ``Panoptic segmentation,'' in \emph{Proceedings of the IEEE Conference on Computer Vision and Pattern Recognition (CVPR)}, 2019.

\bibitem{cas1-mousavian2018joint}
A.~Mousavian, A.~Toshev, and A.~Fathi, ``Joint object and pose recognition using conditional latent-variable models,'' in \emph{Proceedings of the IEEE Conference on Computer Vision and Pattern Recognition}, 2018, pp. 7753--7761.

\bibitem{cas5-zhou2020object}
Y.~Zhou, H.~Zheng, B.~Zhao, and J.~Liu, ``An object recognition and localization method based on category-agnostic instance segmentation,'' in \emph{Proceedings of the 2020 IEEE International Conference on Artificial Intelligence and Computer Applications (ICAICA)}.\hskip 1em plus 0.5em minus 0.4em\relax IEEE, 2020, pp. 51--55.

\bibitem{cas2-li2020real}
Y.~Li, X.~Wang, W.~Cao, J.~Li, and H.~Lu, ``A real-time system for robotic bin picking based on category-agnostic instance segmentation,'' in \emph{2020 IEEE International Conference on Robotics and Automation (ICRA)}.\hskip 1em plus 0.5em minus 0.4em\relax IEEE, 2020, pp. 10\,432--10\,438.

\bibitem{cas3-qin2019efficient}
L.~Qin, S.~Yang, and H.~Liu, ``An efficient and robust category-agnostic instance segmentation method for bin-picking robots,'' \emph{Robotics and Autonomous Systems}, vol. 120, pp. 104--116, 2019.

\bibitem{cas4-yin2021bin}
X.~Yin, Y.~Huang, X.~Li, Y.~Zhang, J.~Li, S.~Li, and S.~Chen, ``Bin-picking for multiple objects based on category-agnostic instance segmentation and hand--eye calibration,'' \emph{IEEE Transactions on Automation Science and Engineering}, vol.~18, no.~3, pp. 1427--1438, 2021.

\bibitem{james2018transferring}
S.~James, A.~J. Davison, and E.~Johns, ``Transferring end-to-end visuomotor control from simulation to real world for a multi-stage task,'' in \emph{Proceedings of the IEEE Conference on Computer Vision and Pattern Recognition (CVPR)}, 2018, pp. 2217--2226.

\bibitem{augmentation2019survey}
C.~Shorten and T.~M. Khoshgoftaar, ``A survey on image data augmentation for deep learning,'' \emph{Journal of Big Data}, vol.~6, no.~1, p.~60, 2019.

\bibitem{james2019sim}
S.~James, A.~Gupta, and S.~Levine, ``Sim-to-real transfer of robotic control with dynamics randomization,'' in \emph{2019 International Conference on Robotics and Automation (ICRA)}.\hskip 1em plus 0.5em minus 0.4em\relax IEEE, 2019, pp. 6914--6920.

\bibitem{navi2018sim}
F.~Sadeghi and S.~Levine, ``Sim2real view invariant visual servoing by recurrent control,'' in \emph{Robotics: Science and Systems (RSS)}, 2018.

\bibitem{loco2018deep}
X.~B. Peng, G.~Berseth, C.~Yin, and S.~Schaal, ``Deep reinforcement learning for robotic locomotion with asynchronous advantage actor-critic (a3c),'' in \emph{IEEE International Conference on Robotics and Automation (ICRA)}.\hskip 1em plus 0.5em minus 0.4em\relax IEEE, 2018, pp. 1--8.

\bibitem{depth_transfer_icra_17}
M.~Johnson-Roberson, C.~Barto, R.~Mehta, S.~N. Sridhar, K.~Rosaen, and R.~Vasudevan, ``Driving in the matrix: Can virtual worlds replace human-generated annotations for real world tasks?'' in \emph{2017 IEEE International Conference on Robotics and Automation (ICRA)}.\hskip 1em plus 0.5em minus 0.4em\relax IEEE, 2017, pp. 746--753.

\bibitem{bin-picking-survey-2020}
M.~Fujita, Y.~Domae, A.~Noda, G.~Garcia~Ricardez, T.~Nagatani, A.~Zeng, S.~Song, A.~Rodriguez, A.~Causo, I.-M. Chen \emph{et~al.}, ``What are the important technologies for bin picking? technology analysis of robots in competitions based on a set of performance metrics,'' \emph{Advanced Robotics}, vol.~34, no. 7-8, pp. 560--574, 2020.

\bibitem{bin-picking-survey-2}
A.~Cordeiro, L.~F. Rocha, C.~Costa, P.~Costa, and M.~F. Silva, ``Bin picking approaches based on deep learning techniques: A state-of-the-art survey,'' in \emph{2022 IEEE International Conference on Autonomous Robot Systems and Competitions (ICARSC)}, 2022, pp. 110--117.

\bibitem{antipodal}
S.~Kumra, S.~Joshi, and F.~Sahin, ``Antipodal robotic grasping using generative residual convolutional neural network,'' in \emph{2020 IEEE/RSJ International Conference on Intelligent Robots and Systems (IROS)}.\hskip 1em plus 0.5em minus 0.4em\relax IEEE, 2020, pp. 9626--9633.

\bibitem{baseline}
S.~V. Pharswan, M.~Vohra, A.~Kumar, and L.~Behera, ``Domain-independent unsupervised detection of grasp regions to grasp novel objects,'' in \emph{2019 IEEE/RSJ International Conference on Intelligent Robots and Systems (IROS)}, 2019, pp. 640--645.

\bibitem{levine2018learning}
S.~Levine, P.~Pastor, A.~Krizhevsky, J.~Ibarz, and D.~Quillen, ``Learning hand-eye coordination for robotic grasping with deep learning and large-scale data collection,'' \emph{The International Journal of Robotics Research}, vol.~37, no. 4-5, pp. 421--436, 2018.

\bibitem{raj2022domain}
P.~Raj, A.~Singhal, V.~Sanap, L.~Behera, and R.~Sinha, ``Domain-independent disperse and pick method for robotic grasping,'' in \emph{2022 International Joint Conference on Neural Networks (IJCNN)}, 2022.

\bibitem{raj_tase23}
P.~Raj, L.~Behera, and T.~Sandhan, ``Scalable and time-efficient bin-picking for unknown objects in dense clutter,'' \emph{IEEE Transactions on Automation Science and Engineering}, pp. 1--13, 2023.

\bibitem{CAD-bin-picking}
L.~D. Hanh and K.~T.~G. Hieu, ``3d matching by combining cad model and computer vision for autonomous bin picking,'' \emph{International Journal on Interactive Design and Manufacturing (IJIDeM)}, vol.~15, pp. 239--247, 2021.

\bibitem{gso}
L.~Downs, A.~Francis, N.~Koenig, B.~Kinman, R.~Hickman, K.~Reymann, T.~B. McHugh, and V.~Vanhoucke, ``Google scanned objects: A high-quality dataset of 3d scanned household items,'' \emph{arXiv preprint arXiv:2204.11918}, 2022.

\bibitem{textures-in-wild}
M.~Cimpoi, S.~Maji, I.~Kokkinos, S.~Mohamed, and A.~Vedaldi, ``Describing textures in the wild,'' in \emph{Proceedings of the IEEE conference on computer vision and pattern recognition}, 2014, pp. 3606--3613.

\bibitem{pytorch}
A.~Paszke, S.~Gross, F.~Massa, A.~Lerer, J.~Bradbury, G.~Chanan, T.~Killeen, Z.~Lin, N.~Gimelshein, L.~Antiga, A.~Desmaison, A.~Kopf, E.~Yang, Z.~DeVito, M.~Raison, A.~Tejani, S.~Chilamkurthy, B.~Steiner, L.~Fang, J.~Bai, and S.~Chintala, ``Pytorch: An imperative style, high-performance deep learning library,'' in \emph{Advances in Neural Information Processing Systems 32}.\hskip 1em plus 0.5em minus 0.4em\relax Curran Associates, Inc., 2019, pp. 8024--8035.

\bibitem{cocodataset}
T.~Lin, M.~Maire, S.~J. Belongie, L.~D. Bourdev, R.~B. Girshick, J.~Hays, P.~Perona, D.~Ramanan, P.~Doll{'{a} }r, and C.~L. Zitnick, ``Microsoft {COCO:} common objects in context,'' \emph{CoRR}, vol. abs/1405.0312, 2014.

\end{thebibliography}


 





\end{document}